# A Simple Data Discretizer

Gourab Mitra
gourab5139014@gmail.com
Dept. of Computer Science and Engineering
SUNY at Buffalo, USA

Shashidhar Sundereisan
shashidhar.sun@gmail.com
Bloomberg L.P., USA

Bikash Kanti Sarkar
bk_sarkarbit@hotmail.com
Dept. of Computer Science and Engineering
Birla Institute of Technology, Mesra, Ranchi, India

**Abstract** Data discretization is an important step in the process of machine learning, since it is easier for classifiers to deal with discrete attributes rather than continuous attributes. Over the years, several methods of performing discretization such as Boolean Reasoning, Equal Frequency Binning, Entropy have been proposed, explored, and implemented. In this article, a simple supervised discretization approach is introduced. The prime goal of MIL is to maximize classification accuracy of classifier, minimizing loss of information while discretization of continuous attributes. The performance of the suggested approach is compared with the supervised discretization algorithm Minimum Information Loss (MIL), using the state-of-the-art rule inductive algorithms- J48 (Java implementation of C4.5 classifier). The presented approach is, indeed, the modified version of MIL. The empirical results show that the modified approach performs better in several cases in comparison to the original MIL algorithm and Minimum Description Length Principle (MDLP).

**Nomenclature**

| | |
|---|---|
| $A_i$ | any continuous attribute of classification problem |
| s | number of class values in classification problem |
| $d_{min}, d_{max}$ | minimum and maximum values of continuous attribute |
| m | total number of examples in a classification problem |
| n | number of sub-intervals of $A_i$ |
| c | user set constant value |
| $S_j$ | j-th sub-interval of $A_i$ |
| h | length of each sub-interval |
| $f_k$ | frequency of examples occurring in the sub-interval $S_j$ |
| TS | expected threshold value for each sub-interval |
| $CTS_j$ | computed threshold value of the j-th sub-interval |
| U | the union operator |
| acc | mean accuracy |

| s.d. | standard deviation |
|------|--------------------|

---

## 1. Introduction

Most of the real-world problems involve *continuous* attributes, each of which possesses more values. Obviously, due to exponential growth of data in the database systems, operating continuous values is comparatively complex in comparison to discrete values. Therefore, conversion of input data sets with continuous attributes into data sets with discrete attributes is necessary to reduce the range of values, and this is indeed known as data discretization. Although, transforming the continuous values of attribute into a few discrete values may be expressed *nominally* (e.g. "low", "medium" and "high" ) or the attribute may even be *nominal* before transformation. The values of nominal attributes are further needed to *normalize* such as: 1(low), 2(medium), 3(high). In Knowledge Discovery of Databases(KDD), discretization process is known to be one of the most important data preprocessing tasks. Discretization of continuous attributes has been extensively studied (Khiops and Boulle, 2004; Catlett,1991; Ching et al.,1995; Chmielewski and Grzymala-Busse, 1996; Dougherty et al., 1995; Fayyad and Irani, 1993; Kohavi and Sahami, 1996; Pal and Biswas, 2005), prior to the learning process.

Several machine learning algorithms have been developed to mine knowledge from real-world problems. However, many of them such as (Apte and Hong, 1996; Cendrowaka,1987; Clark and Niblett T., 1989) cannot handle continuous attributes, whereas each of them can operate on discretized attributes. Furthermore, even if an algorithm can handle continuous attributes its performance can be significantly improved by replacing continuous attributes with its discretized values(Catlett, 1991; Pfahringer,1995).The other *advantages* in operating discretized attributes are the need of less memory space and processing time in comparison to their non-discretized form. In addition, much larger rules are produced, while processing continuous attributes. On the other hand, a *disadvantage* of discretizing a continuous value in some cases, is the loss of the information in the available original data. For example, two different values in the same discretization interval are considered equal, even though they may be at two different extremes of the interval. Such an effect can reduce the precision of Machine Learning algorithm(Ventura and Martinez,1995). The loss of information, in fact, increases the rate of misclassification instead of improvement of classification accuracy.

In machine learning, discretization methods can be divided into number of categories such as: *unsupervised* or *supervised*, *local* or *global*. Unsupervised discretization methods are not provided with class label information, whereas supervised discretization methods are supplied with a class label for each data item value. Again, local methods are restricted to discretizing single continuous attribute per run, whereas global ones operate on more than one continuous attribute at the same run. Equal Frequency Binning(Liu et al., 2002) is a simple unsupervised and local discretization algorithm which discretizes continuous valued attributes by creating a specified number of bins. On the other hand, MDLP ( Fayyad and Irani, 1993) is an entropy based supervised and local discretization method. Further, ChiMerge (Kerber, 1992) and Chi2(Liu and Setiono, 1997) are the local methods that provide statistically justified heuristic method for supervised discretization. In particular, both the heuristic methods discretize attribute by taking account into the interdependence between class labels and attribute values. Recently, Kurgan and Cios(2004) have developed a new supervised approach CAIM(class-attribute independence maximization) to minimize the loss between class-attribute interdependency, finding minimum number of sub-intervals. Interestingly, a brief review on the comparison study among the discretization approaches is presented below.

Dougherty et al. (1995) compare discretization results obtained by unsupervised discretization versus a supervised method presented by Holte (1993) and the entropy based method proposed by Pfahringer (1995). They conclude that supervised methods are better than unsupervised discretization method as they generate fewer classification errors. Kohavi and Sahami (1996) show that the number of classification errors generated by the discretization method of Fayyad and Irani (1993) is comparatively smaller than the number of errors generated by the discretization algorithm proposed by Auer et al. (1995). They recommend that entropy based discretization methods are usually better than other supervised discretization algorithms. Recently, Ruoming et al. (2009) have reported that their proposed dynamic programming approach significantly reduces the classification errors as compared to the existing discretization approaches. However, it is well accepted that there is no superior discretization method that gives best results across all domains. Marzuki and Ahmad (2007) have suggested that different discretization methods perform better in one or more domains.

In general, most of the discretization processes lead to a loss of information and can't reduce values of discrete attribute. The MIL discretizer (Sarkar et al., 2011) has much emphasized in this respect. However, there are scopes to improve this algorithm, and the present study has focused on these points. In particular, the loss of information is, in fact, measured by the percentage of mis-classification (i.e., error) on unknown examples. Further, the missing values and the non-numeric values of the attributes are here handled suitably. The performance of the approach is compared with the supervised MIL, using rule inductive algorithms- J48 (Waikato Environment for Knowledge Analysis -3.4.2). The experimental results presented in this article demonstrates that the present version of MIL algorithm favorably improves in several cases as compared to the original MIL and MDLP (WEKA-3.4.2).

This paper begins with discussing the importance of discretization method in data mining. In Section-2, the proposed approach modified MIL is presented in details, following a brief on simple discretization process. Section-3 shows the experimental results. Finally, conclusion is summarized in Section-4.

## 2. Discretization by Modified Minimum Information Loss (MIL) Discretizer

*Brief idea on simple discretization process***:**

In particular, a discretization problem takes a set of instances as input, each of which has values for a set of attributes. The set of attributes can contain all discrete attributes or all continuous attributes or a combination of discrete and continuous attributes. Optionally, one of these can be a special attribute called class attribute. On the other hand, a discretization process converts input data sets with continuous attributes (if any) into data sets with discrete attributes. However, each discretization technique must be such that the accuracy of the prediction made from the discretized data is at least as good as the prediction made from the original continuous data. In practice, a simple discretization process consists of the following four major steps:

(i) sort all the continuous values of the attribute to be discretized,
(ii) choose a cut point to split the continuous values into sub-intervals,
(iii) split or merge the sub-intervals of continuous values,
(iv) choose the stopping criteria of the discretization process.

*Original MIL approach:*

Ideally, the goal of the proposed discretization approach MIL is to achieve maximum performance in terms of classification accuracy while minimizing the loss of information (*i.e.*, mis-classification rate). The approach follows the *split and merge* strategy. In other way, a set of sub-intervals for each continuous attribute is initially decided and these are merged afterward. However, for optimal merging, two parameters namely, *threshold value* (TS) and *computed threshold value* (CTS) for each sub-interval play an important role. The parameters are defined below.

Suppose a continuous attribute $A_i$ of a classification problem with *m* examples possesses $\ell$ and *u* as the *lower* and *upper* limits respectively. Now, assume that the entire range of any continuous attribute is initially split into *n* sub-intervals, where *n* (n < m) is decided as:   n = c∗s  ------------ (1)

Here, '*c*' is an user set *constant*, and '*s*' represents the number of values of the *class* (target) attribute in the same classification problem. In particular, *n* is chosen as proportional to the number of class values, assuming that attribute information disperses over the data set as per the values of class attribute. Clearly, total number of selected points (*i.e.*, values) over the entire range must be n+1, since the number of sub-intervals is *n*. The points are denoted here as: $d_i$, $0 \leq i \leq n$.

Now, assuming *uniform* distribution of data over the entire range of the attribute, the TS of each sub-interval is decided as:

$$TS = \frac{m}{n} \cdots\cdots\cdots\cdots\cdots (2)$$

It is, in fact, the *expected frequency* of each sub-interval. Of course, such a uniform distribution for each sub-interval does not practically occur. For this reason, interest is given to measure CTS of each sub-interval with the aim to merge sub-intervals. The CTS of each sub-interval simply tells the *number* of samples practically occurring within itself. So, $CTS_j$ of any sub-interval $S_j$ ($1 \leq j \leq n$), can be expressed as follows:

$CTS_j = f_k \cdots\cdots\cdots\cdots (3)$, where $f_k$ is the frequency of examples occurring in $S_j$.

**The MIL algorithm:**

In fact, the MIL algorithm starts with a set of sub-intervals of an attribute, and finally results a minimal number of merged sub-intervals. More specifically, the algorithm uses a greedy approach which searches for promising optimal merged sub-intervals (each represents a discrete value) as per the criteria specified at Step-3. Roughly speaking, the algorithm consists of two basic steps (given below).

- Initialization of discretization scheme
- Consecutive addition of sub-intervals in order to result optimal merged sub-intervals (i.e., regions) based on the MIL criterion.

The pseudo-code of the approach is presented below.

*Assumptions*:
- $x \in [d_l, d_u)$ implies $d_l \leq x < d_u$.
- $[d_j, d_{j+1}]$ represents the j-th sub-interval with *lower* and *upper* limits: $d_j$ and $d_{j+1}$ respectively.
- $[d_j, d_r]$ represents a merged sub-interval with *lower* and *upper* limits: $d_j$ and $d_r$ respectively.
- The term *region* means collection of one or more consecutive sub-intervals by merging (i.e., it is a merged sub-interval).
- CTS(R) implies CTS of region R.

*Input*: Data set consisting of *m* examples, *s* classes, and *c* (user set constant value)
*Variables*:
   j, r         // index variables, and    $0 \leq j < r \leq n$
   Ctr        // counter for checking the status of number of the processed sub-intervals
   NS        // counter to maintain the status of number of optimal merged sub-intervals
   MPVal    // to maintain the discrete value to each resultant optimal merged sub-interval
   Tot_CTS     // total CTS value of the current region
                    // Tot_CTS is used for simpler implementation

   *for* every continuous attribute $A_i$ *do*
Step-1:
  1.1. Find the minimum ($d_{min}$) and the maximum ($d_{max}$) values of $A_i$.
  1.2. Initialize Ctr = 0, NS = 0, MPVal = 0, j = 0
  1.3. Compute total number of sub-intervals: $n = c*s$, length of each sub-interval
       $h = \frac{d_{max} - d_{min}}{n}$ and $TS = \frac{m}{n}$.

Step-2:
  2.1. Replace each missing value (if any) of $A_i$ in example(s) by a random number lying in $(d_{min}, d_{max})$.
  2.2. Find $CTS_i$ ($1 \leq i \leq n$), (i.e., CTS of each sub-interval with length h, starting at the point $d_{min}$), Tot_CTS = $CTS_1$ (i.e., CTS of the 1$^{st}$ sub-interval R:[$d_0, d_1$] )

Step-3:
  3.1. *If* CTS(R) (i.e., CTS of the current region R: [$d_j, d_r$), $j+1 \leq r < n$) is less than $\left\lceil \frac{TS}{3} \right\rceil$ of the same region R, *then* do the following sub-tasks.

    3.1.1.   TS = TS + $\frac{m}{n}$.
    3.1.2 Update Tot_CTS by adding CTS(r) of the next sub-interval: [$d_r, d_{r+1}$), i.e., Tot_CTS = Tot_CTS + CTS(r)
    3.1.3 Merge R with [$d_r, d_{r+1}$), i.e., R= R U [$d_r, d_{r+1}$), and set r = r+1
             CTS(R) = Tot_CTS

  *else* do the following sub-tasks: // i.e., *when an **optimal** sub-interval is found*
   3.1.4. j = r, NS = NS +1, MPVal = MPVal + 1, set the current MPVal to the region R.
   3.1.5. Set $TS = \frac{m}{n}$,   R = [$d_j$, $d_{j+1}$),   Tot_CTS = CTS(R)
 3.2. Set Ctr = Ctr +1.
Step-4: Continue Step-3 until (Ctr ≠ n).
  // i.e., *until n sub-intervals of the attribute $A_i$ are processed*
Step-5: Set the current MPVal to the last region, say [$d_j$, …), where '…' in [$d_j$, …) implies all values > $d_j$.
Step-6: Assign a discrete value for each continuous value of the attribute as per the respective merged sub-interval's range.
<u>*Output:*</u>   A set of discrete values (each associates with an *optimal* merged sub-interval represented by MPVAL), and simultaneously discretization of continuous values of the attribute (as per the discrete values assigned for the respective merged sub-intervals).

***Discussion on the algorithm:***
Clearly, the presented algorithm starts with a number of sub-intervals initially set *n* (depending on the values of class attribute), and emphasizes to return *minimum* number of merged sub-intervals as the *outcome* for each continuous attribute. In other way, it attempts to return minimum value for the variable MPVal (representing discrete values). Further, the algorithm terminates in *linear* time, since the continuous attribute need not be *sorted*. In particular, to generate discrete form of non-discretized attribute, only *four* consecutive *scans* for the values of the attribute are necessary. The *first* scan finds the minimum ($d_{min}$) as well as the maximum ($d_{max}$) values of the continuous attribute, number of sub-intervals and length of each sub-interval. All these are performed by Step-1. On the other hand, the *next* scan replaces missing value of attribute, and simultaneously finds CTS value of each sub-interval (starting at $d_{min}$). These are done by Step-2. Note that any missing value is replaced by a value (represented by *replace_val*) which is found by the formula:

$replace\_val = d_{min} + \frac{d_{max} - d_{min}}{m} \times g\_val$, where *g-val* represents a random value

lying in (1, *m*). The *third scan* decides the optimal merged sub-intervals. As soon as an optimal merged sub-interval (say R) is found, values of NS and MPVal increase, and the approach reinitializes the values of the parameters, namely TS, Tot_CTS and R to search for the next merged sub-interval. All these operations are, indeed, performed by Step-3 of the algorithm. Obviously, Step-3 continues until

*Ctr* (representing the number of sub-intervals) reaches to *n*. However, when Step-3 ends, the latest value of MPVal is set for the last region, say $[d_j, …)$, where $[d_j , …)$ implies all values $\geq d_j$. These are done through Steps 4 to 5 of the algorithm. The *final scan* (performed by Step-6) replaces each continuous value of attribute with the discrete value (assigned to the respective merged sub-interval). For deeper understanding the algorithm, one may look up at *Appendix*- III(a) of this chapter.

Now, it is necessary to point out that the values of the following two parameters are here decided based on *trial* and *error* method.

i) The value of constant *c* used in *equation* (1).
ii) The value of *k* in $\left\lceil \frac{TS}{k} \right\rceil$ (used in *Step*-3.1 of the algorithm) for searching the lower and the upper bounds of the *optimal* merged intervals

Although, it is observed through experiments that if the values: 20 and 3 are set respectively to *c* and *k*, then, on an average, better classification accuracy results for data sets are achieved. Moreover, the pair of values: (20, 3) is capable to reduce number of discrete values of attributes as well as the *error-rate* over data sets. As evidences, the results of two experiments over *Heart* (Swiss) data set are shown in Appendix-III(b). This data set is chosen because the performance of any classifier over this data set is not good enough.

Let us consider also an analytical discussion in this regard. *For instance*, if we set criteria like (TS/2), instead of $\left\lceil \frac{TS}{3} \right\rceil$ in Step-3.1 of the algorithm, then the probability of decreasing number of discrete values may be high, and that may simultaneously loose information of the attribute. On the other hand, criteria like (TS/4) instead of $\left\lceil \frac{TS}{3} \right\rceil$, set in Step-3.1 increases the probability of resulting more number of discrete values (*i.e.*, almost closer to original range of attribute). Accordingly, saving space and time due to discretization is not expected up to mark.

***Analysis on time complexity:***
If the data set contains total *m* samples, then the MIL algorithm takes only O(*m*) (i.e., linear) time to discretize each continuous attribute, since it needs only *four* scans to discretize a continuous attribute. Although, it is safe to say that *space* complexity of MIL algorithm is comparatively more in comparison to any state of art discretizer like MDLP. It is approximately $O(n)$, since two extra arrays (each of size $O(n)$) are used to find TS and CTS for $n$ initial sub-intervals.

Thus, the complexity analysis (mainly time complexity) recommends that MIL algorithm is well suited for large problems too.

**Modified MIL approach:**

The MIL algorithm discretizes by merging only those intervals which had very few members (i.e., CTS(R) < TS/3). However, the present opinion on merging of these intervals would better represent the continuous data if these intervals have similar number of members in them, as because the earlier version has not concentrated on those intervals, each of which have members more than TS/3. A brief outline especially focusing on the modification of the original MIL algorithm is presented below. Actually, the modification attempts to cater to this observation.

1. Find the minimum ($d_{min}$) and the maximum ($d_{max}$) values of Attribute $A_i$.
2. $n = c*s$, where $n$ = initial no. of intervals, $c$ = a constant chosen by the user and $s$ = no. of classes in the target attribute.
3. Replace each missing value (if any) of $A_i$ in example(s) by a random number lying in ($d_{min}$, $d_{max}$).
4. $h = (d_{max} - d_{min})/n$, where $h$ = length of each sub-interval.
5. Divide the range into n intervals the ith interval having the range [$d_{min} + i*h, d_{min} + (i+1)*h$) where i belongs to [1,n].
6. Define a new array CTS[1…n] which stores the no of members in each sub interval defined as in step 5.
7. TS=m/n, where m = no. of instances in training data set.
8. Initialize i = 1, TOT_CTS=CTS[0], last_small_merge = false.
9. *If* TOT_CTS < TS/3, *then*
    9.1.  begin
    9.2.  Merge the current region R with interval $d_{i+1}$ i.e. R = R U $d_{i+1}$.
    9.3.  Update TOT_CTS = TOT_CTS+CTS[i+1]
    9.4.  TS=TS + m/n, i = i + 1.
    9.5.  last_small_merge = true.
    9.6.  end.

Step 10:    Else
    10.1.    begin
    10.2.    If(0.75*CTS[i+1]≤CTS[i]≤1.25*CTS[i+1] and last_small_merge = false )
        10.2.1. begin
        10.2.2. Merge the current region R with interval $d_{i+1}$ i.e. R = R U $d_{i+1}$.
        10.2.3. update i = i + 1. TOT_CTS = TOT_CTS+CTS[i+1]
        10.2.4. end.
    10.3.    else
        10.3.1 begin
        10.3.2.    Update TOT_CTS = CTS[i+1], TS = m / n,
            last_small_merge = false.
        10.3.3. end
    10.4   end
Step 11:  If (i < n) goto Step 9.
Step 12: Assign a discrete value for each continuous value of the attribute as per the respective merged sub-interval's range.

In short, intervals are either merged or left alone. If they are merged then they are merged according to the rules:
   i)   TOT_CTS < TS/3 or
   ii)  0.75*CTS[i+1] ≤ CTS[i] ≤ 1.25*CTS[i+1].

After the intervals are merged, each of them is discretized by taking the mean of the new ranges. For Each instance in the training data set the continuous value is replaced with the discretized value. This is done for all the attributes in the data that need discretization.

## 3. Experimental Design and Results

In this section, the details of experiments carried out are first described. Then, the accuracy results on the selected data sets are reported and analyzed appropriately. No doubt, all experiments in this study are performed on the same machine.

### *Experimental Design*:

For empirical evaluation of the discretizers: MIL and the modified-MIL, *eleven* well-known continuous and mixed-mode data sets are collected from UCI machine learning repository (Blake et al., 1999). A summary of the data sets is given in Table-1. The table describes the total number of *attributes*, the number of *continuous* attributes, the number of *classes* and the *size* of each data set. Note that the presented algorithm and the original MIL algorithm are implemented in Java-1.4.1 on a Pentium-4 m/c running Mandrake Linux OS 9.1, whereas MDLP is under WEKA-3.4.2 frame work.

For empirical evaluation of MIL, modified-MIL and MDLP , an well known state-of-the-art classification algorithm: the J48 (Java implementation of *C4.5 decision tree*) is chosen. The classifier is tested on each data set but descretized by namely MIL, modified-MIL and MDLP discretizers .

To quantify the impact of the performance of discretization algorithms, results (i.e., classification accuracies) achieved by the classifier on each discretized data set over 50 runs are averaged. Also, the computed standard deviation(s.d) on the basis of these 50 results is recorded along with the respective *mean* value. However, in each of 50 runs, 30% of the total data is selected randomly as *training* set and the rest as *test* set, maintaining almost uniform class distribution in each training set. Obviously, the main intention behind selecting less data for training set is to reduce the biasness of classifier towards training data.

The *classification accuracy* of a rule set generated by any classifier is expressed as a *percentage* of unseen data correctly classified and defined as:

$$\text{Accuracy} = \frac{\text{No. of test examples correctcly classified by the rule set}}{\text{Total number of test examples}} \times 100 \quad \ldots\ldots(4)$$

Table-1: Description of the UCI data sets

| Problem Name | Number of non-target attributes | Number of continuous attributes | Number of classes | Number of examples |
|---|---|---|---|---|
| Adult | 14 | 06 | 02 | 48842 |
| Ecoli | 7 | 7 | 8 | 336 |
| Glass | 9 | 9 | 6 | 214 |
| Haberman | 3 | 3 | 2 | 306 |
| Heart (Hungarian) | 13 | 5 | 5 | 294 |
| Heart (Swiss) | 13 | 4 | 5 | 123 |
| Iris | 4 | 4 | 3 | 150 |
| Letter-recognition | 16 | 16 | 26 | 20000 |
| TAE | 5 | 5 | 3 | 151 |
| Transfusion | 4 | 4 | 2 | 748 |
| Vertebral | 6 | 6 | 2 | 310 |

Table-2: Accuracy(%) comparison using J48 on the data sets
( acc = average accuracy, s.d .= standard deviation)

| Problem Name | Modified-MIL Discretizer ( acc ± s.d. ) | MIL Discretizer ( acc ± s.d. ) | MDLP Discretizer ( acc ± s.d. ) |
|---|---|---|---|
| Adult | **84.62** ± 4.12 | 84.27 ± 4.15 | 86.21 ± 4.34 |
| Ecoli | 90.54 ± 9.14 | 90.84 ± 9.13 | 90.49 ± 9.16 |
| Glass | **80.66** ± 5.83 | 79.13 ± 6.18 | 81.76 ± 5.37 |
| Haberman | 75.16 ± 7.98 | 76.79 ± 7.65 | 70.81 ± 9.23 |
| Heart (Hungarian) | **72.32** ± 5.22 | 71.18 ± 5.45 | 70.98 ± 6.52 |
| Heart | **49.82** ± 7.35 | 47.34 ± 8.12 | 45.38 ± 9.43 |
| (Swiss) | | | |
| Iris | 98.78 ± 3.82 | 98.78 ± 3.82 | 97.03 ± 3.78 |
| Letter-recognization | 87.95 ± 4.37 | 87.95 ± 4.37 | 78.85 ± 6.23 |
| TAE | **62.96** ± 6.13 | 63.32 ± 6.04 | 62.18 ± 6.52 |
| Transfusion | **77.40** ± 4.12 | 77.27 ± 4.15 | 76.20 ± 4.32 |
| Vertebral | **85.16** ± 6.05 | 81.93 ± 7.83 | 84.14 ± 6.52 |

The above performance table shows improved results in favour of modified-MIL discretization on seven datasets, namely *Adult*, *Glass*, *Heart* (Hungarian), *Heart* (Swiss), *TAE*, *Transfusion* and *Vertebal* in comparison to original MIL. On the other hand, the results of MDLP, MIL and the modified-MIL are almost same on the rest.

## 4. Conclusion and Future Work

In the present investigation, the initial number of sub-intervals for each continuous attribute is user set. This number is, in fact, decided here as *proportional* to the number of class values in the classification problem, assuming that variation of attribute information disperses throughout the entire range according to the *variety* of class values. Definitely, if the number of class values increases, then the number of sub-intervals should increase. Consequently, such a decision hopes to reduce loss of information, while discretization. In this connection, it is observed based on *trial and error* method that if the value of constant $c$ is initially set 20, then the performance of discretized data set is, on an average, better as compared to the other values of $c$. Although identifying the best value of $c$ is a challenging task, as it cannot be universally applied to each attribute of a problem.

The classification results using discretized data show that the modified-MIL algorithm in comparison to the other two leading algorithms improves classification accuracies achieved by the subsequently used machine learning algorithms. Further, the computed *standard deviation* values of the classification accuracies conclude that the performance of modified version is more reliable for classification.

For the safe, it may be pointed out that if CTS value of each sub-interval for any continuous attribute is greater than or equal to its TS, then there exists a probability of generating $n$ discrete values (*i.e.*, equal to initial number of sub-intervals) by this algorithm. Similarly, only one discrete value may be generated if density of examples occurs only at the last sub-interval. However, these two cases are practically rare.

In a nutshell, the goodness of the present approach MIL is evaluated based on *missing* value handling criteria for attributes, the number of generated *discrete values* of attributes, the classification *accuracy* and the *execution time*, capability of handling *volumetric* data.

There is scope for further improvement in MIL(i.e., deciding number of distinct sub-intervals for each continuous attribute).

# Appendix- 1

## a) Illustration of original MIL algorithm:

For better understanding the algorithm MIL, let us present some values of a continuous attribute A of a problem with 240 examples.

```
A       ……………………
0.50
0.10
0.20
0.15
0.25
0.31
0.71
0.52
0.82
0.131
0.90
0.12
…..
…..
```

Clearly, m=240. Suppose that n(number of sub-intervals)=40, $d_{min}$= 0.1 and $d_{max}$=0.90.

Now, TS= (m/n)= (240/40) =6, h = ($d_{max}$- $d_{min}$)/n =(0.9 – 0.1)/40 = 0.8/40 = 0.020.

Thus, $I_1$( first sub-interval) = [0.10, 0.120), $I_2$(second sub-interval) = [0.120, 0.140), and so on. *All the above computations are performed through step-1 to step-2 of the presented algorithm.*

Assume that CTS of $I_1$ is less than $\frac{TS}{3}$ = 2, so merge $I_1$ with the next sub-interval $I_2$.

Now, assume that the sum of CTS of $I_1$ and $I_2$ is greater than or equal to $2*\frac{TS}{3}$.

So, the first merged optimal sub-interval is found, and it is $R_1$ = [0.10, 0.140). and hence MPval = MPval + 1 = 0 + 1 = 1 is set to $R_1$ ( i.e., it is the *first* found discrete value ).

*All the above computations are performed by step-3 of the presented algorithm.*

Step-3 continues until n sub-intervals are processed. At step-5, the respective discrete value is set to the last region, i.e., a region R consisting of the value ≥ 0.90.

The final step of the algorithm, i.e., step-6 assigns discrete values to the continuous values of the attribute. For example,

```
A       ………………………
0.50
0.10        → 1
0.20
0.15
0.25
0.31
0.71
0.52
0.82
0.131       → 1
0.90
```

|        |     |
|--------|-----|
| 0.12   | → 1 |
| …..    |     |

In the same fashion, other continuous values get simultaneously the respective discrete values.

**a) *Illustration of the modified-MIL algorithm*:**

For better understanding of the modified mil algorithm consider an example with 100 instances. So m = 240. Lets assume there are 2 classes i.e. s = 2. Let the user specified constant c be 20. Thus n = c*s i.e. 40.

TS = m/n which is 6.

Let the array attribute be
A ………………………
0.50
0.10
0.20
0.15
0.25
0.31
0.71
0.52
0.18
0.131
0.90
0.12
…..

Let $d_{min}$= 0.1 and $d_{max}$=0.90 .
So h = ($d_{max}$- $d_{min}$)/n =(0.9 – 0.1)/40 = 0.8/40 = 0.020.

Thus, $I_1$( first sub-interval) = [0.10, 0.120), $I_2$(second sub-interval) = [0.120, 0.140), and so on.

Now the algorithm is run up to Step 9.

Say CTS(1) = 1. Since it is less than TS/3(2) We merge it with the next interval. Now, assume that the sum of CTS of $I_1$ and $I_2$ is greater than or equal to 2* $\frac{TS}{3}$. So, the first merged optimal sub-interval is found, and it is $R_1$ = [0.10, 0.140). This step is the same as the MIL algorithm.

If CTS[3] = 4 and CTS[4]= 5. Since CTS[4] <=1.25*CTS[3]. We merge the two intervals to form R = (0.140,0.180]. This step is the main difference between the two algorithms.

These steps are repeated until i = n.
The step 12 of the algorithm just assigns discrete values to all the values of the attribute.

```
    A   ………………………
   0.50
   0.10  → 1
   0.20
   0.15  → 2
   0.25
   0.31
   0.71
   0.52
   0.18 → 2
   0.131  → 1
   0.90
   0.12  → 1
   ……
```

**Note**  However, the set of discrete values (each associates with an *optimal* merged sub-interval) and simultaneously discretization of continuous values of the attributes as per the merged sub-interval's range (obtained in the output step of the MIL algorithm) may be recorded properly in a file to discretized any unseen example of the same classification problem in context of prediction.

Further, any value lesser than the minimum value($d_{min}$) of a continuous attribute(found in future in an example) of a classification problem may be treated under the first interval(found by MIL) for the same attribute.